\documentclass[letterpaper, 10 pt, conference]{ieeeconf}  % Comment this line out if you need a4paper
\usepackage{cite}
\usepackage{amsmath,amssymb,amsfonts}
\usepackage{algorithmic}
\usepackage{subcaption}
\usepackage{graphicx}
\usepackage{textcomp}
\usepackage{xcolor}
\let\labelindent\relax
\usepackage{enumitem}
\usepackage{placeins}
\usepackage[ruled,vlined]{algorithm2e}
\usepackage{xcolor}
\def\BibTeX{{\rm B\kern-.05em{\sc i\kern-.025em b}\kern-.08em
    T\kern-.1667em\lower.7ex\hbox{E}\kern-.125emX}}

\IEEEoverridecommandlockouts                              % This command is only needed if 
\begin{document}

\title{\LARGE \bf
Approximating a deep reinforcement learning docking agent using linear model trees
}

\author{Vilde B. Gjærum$^{1}$ and Ella-Lovise H. Rørvik$^{2}$ and Anastasios M. Lekkas$^{1}$\thanks{This work was supported by the Research Council of Norway through the EXAIGON project, project number 304843. An additional thanks to Nicolas B. Carbone for his contribution through several valuable discussions.}\thanks{$^{1}$Department of Engineering Cybernetics, Norwegian University of Science and Technology, Trondheim, Norway. Email: {vilde.gjarum, anastasios.lekkas\}@ntnu.no}}\thanks{$^{2}$Department of Artificial Intelligence, TrønderEnergi, Trondheim, Norway. Email:elh.rorvik@gmail.com}}

\maketitle

%%%%%%%%%%%%%%%%%%%%%%%%%%%%%%%%%%%%%%%%%%%%%%%%%%%%%%%%%%%%%%%%%%%%%%%%%%%%%%%%

\begin{abstract}

Deep reinforcement learning has led to numerous notable results in robotics. However, deep neural networks (DNNs) are unintuitive, which makes it difficult to understand their predictions and strongly limits their potential for real-world applications due to economic, safety, and assurance reasons. To remedy this problem, a number of explainable AI methods have been presented, such as SHAP and LIME, but these can be either be too costly to be used in real-time robotic applications or provide only local explanations. In this paper, the main contribution is the use of a linear model tree (LMT)  to approximate a DNN policy, originally trained via proximal policy optimization(PPO), for an autonomous surface vehicle with five control inputs performing a docking operation. The two main benefits of the proposed approach are: a) LMTs are transparent which makes it possible to associate directly the outputs (control actions, in our case) with specific values of the input features, b) LMTs are computationally efficient and can provide information in real-time. In our simulations, the opaque DNN policy controls the vehicle and the LMT runs in parallel to provide explanations in the form of feature attributions. Our results indicate that LMTs can be a useful component within digital assurance frameworks for autonomous ships. 
\end{abstract}

%%%%%%%%%%%%%%%%%%%%%%%%%%%%%%%%%%%%%%%%%%%%%%%%%%%%%%%%%%%%%%%%%%%%%%%%%%%%%%%%

\section{Introduction}

Deep reinforcement learning (DRL) is a powerful tool with many application areas within robotics, such as perception and control. One of DRL's attributes is that it enables end-to-end learning, which refers to mapping sensory input directly to control actions. This mapping allows for optimizing the overall system performance, instead of having several, locally optimized systems in cascade, which often is the case for model-based systems. In \cite{SinghAvi2019ERRL}, the learned policy was able to perform various manipulation tasks with a dexterous, robotic hand. In \cite{HaarnojaTuomas2018LtWv}, a real-world Minitaur robot learned to walk on a flat surface and was able to handle somewhat challenging surfaces and obstacles without having seen these obstacles during training. In \cite{BakerBowen2019ETUF}, one of DRL's greatest strengths is demonstrated, namely discovering strategies through the exploration of the state- and action space in the multi-agent environment of playing hide and seek. The agents adapted and came up with new strategies to combat the opponents' latest strategy, even going as far as using parts of the environment in ways not originally intended. The applicability of DRL has also been demonstrated in motion control tasks for autonomous surface vessels (ASVs), which often operate in complex and uncertain environments that are challenging to model. In \cite{MartinsenLekkas}, DRL was used to perform curved-path following on a surface vessel and performed well compared to line-of-sight guidance in simulations. In \cite{MeyerHeibergRasheedSan}, a DRL-agent was trained to perform both path-following and collision avoidance. In \cite{RorvikElla-LoviseHammervold2020Adae}, a DRL agent is trained to perform the approach and berthing phases of docking of an ASV.

Even though DRL is a promising tool for advancing the level of autonomy in many fields, its potential applications in real life are strongly limited by the lack of transparency of the deep neural networks (DNNs) involved. This is crucial in all cost- and safety- critical applications. To be able to understand the agent's actions, a \textit{global explainer} is needed, or as a bare minimum, \textit{local explanations} for each prediction. There has been done a lot of work on addressing this problem in the field of eXplainable Artificial Intelligence (XAI) in the recent years. The goal of XAI is to uncover the inner workings of black-box models. One of the most prominent explainers is the Local Interpretable Model-agnostic Explainer (LIME)  \cite{RibeiroMarcoTulio2016WSIT}, which trains an interpretable, surrogate model around the instance it is explaining based on neighboring instances. LIME is a post-hoc (it explains previously trained methods), model-agnostic (it can explain any type of model) XAI method.
% datapoints vs instances vs predictions
% LIME more clear 
The neighboring datapoints are weighted based on how far away from the instance to be explained they are. One weakness of LIME is that it does not perform as well when explaining instances it has not seen before. This problem is addressed in \cite{anchors:aaai18} by the same authors, where the interpretable surrogate model is replaced by a set of IF-THEN rules called \textit{anchors}. Another prominent XAI method is Shapley Additive exPlanations (SHAP) from \cite{LundbergScott2017AUAt}. The  SHAP method explains a prediction by assigning importance to the input features for that prediction. The importance of the features is calculated by utilizing \textit{Shapley values} from game theory, in combination with the coefficients of a local linear regression. SHAP is a model-agnostic, post-hoc method. The assigned contributions of the input features should add up to the original prediction, thus SHAP is an \textit{additive feature attribution} method. Also, although SHAP is mainly a local explanation method, it can give indications of how the black-box model works as a whole through calculating the Shapley values for every instance and analyzing the resulting matrix of Shapley values. It should be noted that SHAP is a very computationally demanding method. Both LIME and SHAP form their explanations by perturbating the inputs and computing how these perturbations affect the output of model to be explained.
In \cite{SlackDylan2020FLaS}, it was shown that XAI methods relying on input perturbations are vulnerable to adversarial attacks aiming to hide their classification bias from the XAI method. One of the reasons such methods are vulnerable to adversarial attacks is that the data sampled from input pertubation often are irrelevant, and the model is forced to explain input samples it has never seen before \cite{KumarI.Elizabeth2020PwSe}. Even if the model to be explained does not intend to fool the explainer, if the samples created by perturbating the inputs are unrealistic, the explanations will be based upon predictions not fairly representing the model. Additionally, in \cite{KumarI.Elizabeth2020PwSe} it is pointed out that SHAP assumes complete independence between the input features, which is often not the case for robotic systems. In \cite{SundararajanMukund2017AAfD}, the method called Integrated Gradients was presented. As the name implies, the gradients are integrated along a straight-line path between the instance to be explained and an information-less baseline instance to extract the explanations directly from the neural network. Integrated Gradients is a post-hoc, model-specific (it can only explain one type of model). 
In this paper, the focus is on linear model trees, a type of decision tree (DT). The rule-based nature of DTs make them inherently transparent and interpretable, since it is trivial to follow the path from the leaf node (output) to the root node (input) of the tree. The most basic form of a decision tree for continuous data - a simple regression tree - has univariate splits and each leaf node predicts a constant value. Model trees are regression trees with a different type of prediction model at the leaf nodes. In linear model trees (LMTs), linear regression is used in the leaf nodes in stead, which makes it easy to extract explanations of the predictions in the form of feature attributions, in addition to being transparent. Linear model trees, as presented in this paper, are fast enough to be used in real-time, are model-agnostic, and can be used to understand both individual predictions and the system as a whole. To the authors' best knowledge, there is no existing literature where LMTs are used to approximate DNN policies controlling robotic systems. The main contributions of this paper are: \begin{itemize}
    \item We use an LMT to approximate a DRL policy, previously trained in \cite{RorvikElla-LoviseHammervold2020Adae} via proximal policy optimization(PPO), to perform autonomous docking in a simulated environment. 
    \item Compared to the standard way of building LMTs, we have added randomization to the search for thresholds and the process of choosing which node to split next. Moreover, to ensure a sufficient dataset from the areas of interest, an iterative approach to the training and data collection was used.
    \item We run the LMT in parallel with the policy in order to provide real-time correlation between input features and the selected control inputs computed by the policy.

%the rest of the apper is structured as follows
\end{itemize}

\section{Docking as a deep reinforcement learning problem}
In this section, the docking problem and the reinforcement learning docking agent are briefly introduced. For further details about the implementation and training, the reader is referred to \cite{RorvikElla-LoviseHammervold2020Adae}. 

\subsection{The docking problem}
Docking pertains to reaching a fixed location along a quay, where the vessel can moor, and can be split in three stages: 1) Moving from open seas to confined waters (\textit{the approach phase}), 2)  parking the vessel (\textit{the berthing phase}), and 3) fastening the vessel to the dock (\textit{the mooring phase}). Docking is a complex motion control scenario and requires a lot of intricate maneuvering, since the vessel operates close to the harbor infrastructure with little to no space for deviations, and under the influence of external disturbances that gain increased importance at lower speeds. Such circumstances are challenging for most traditional control systems since they often depend on accurate mathematical models in contrast to RL-methods that can learn the model guided by the reward function. In \cite{RorvikElla-LoviseHammervold2020Adae}, a PPO policy was trained in a simulated environment based on the Trondheim harbor environment.

\subsection{The docking agent}
\begin{figure}[tb]
    \centering
\includegraphics[trim={0 0 0 1cm},width=0.4\textwidth]{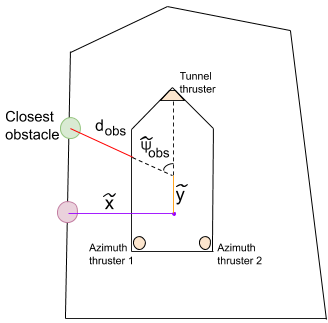}
    \caption{The features, or states, representing the vessels position relative to the closest obstacle and the positioning of the vessels three thrusters.}
    \label{fig:env_ill}
\end{figure}

Deep RL is a subfield of machine learning where learning occurs by selecting actions via an exploration/exploitation scheme and receiving reinforcement signals for these actions. The reinforcement signals, called rewards, are given by the reward function, which is user-defined. The agent is tasked to find the state-action mapping (i.e. the policy) that optimizes the $return$, which represents the expected cumulative reward during an episode. Thus the reward function is crucial for the agent's learning process and its resulting behavior. The policy used in this work, was trained extensively in \cite{RorvikElla-LoviseHammervold2020Adae} with the PPO method from \cite{SchulmanJohn2017PPOA}. It performs the approach and berthing phases of the docking process from up to 400 meters distance from the quay. The PPO method uses a trust region to prevent the agent from overreacting to a training batch and thus risking getting stuck in a local minimum. A trust region is defined as the region where the policy approximation used for gradient descent is adequately accurate. Instead of having the trust-region as a hard constraint, PPO includes it in the objective as a penalty for leaving the trust region, which makes the training less rigorous. The policy is trained to perform the approach- and berthing phase of the docking. The training algorithm has no prior knowledge of the inner dynamics of the vessel and it utilizes the  feedback from the reward function as the agent takes action and the outcomes of these actions are evaluated. Selecting the input features vector is crucial for the reward function and the agents learning, and thorough work was done in \cite{RorvikElla-LoviseHammervold2020Adae}  to develop an effective reward function for the task in hand. The fully-actuated vessel to be controlled has two azimuth thrusters and one tunnel thruster, hence giving the following control states:

\begin{equation}\label{eq:action_vector}
    \textbf{A} = [f_1,f_2,f_3, \alpha_1, \alpha_2],
\end{equation}

where $f_1$,$f_2$ (from -70 to 100 kN) and $\alpha_1$,$\alpha_2$ (from -90 to 90 degrees) are the forces and rotation angles of the two azimuth thrusters, whereas the tunnel thruster can exert only a lateral force $f_3$ (from -50 to 50 kN). The thrusters' placement on the vessel can be seen in Figure \ref{fig:env_ill}. The state vector, which is also the input feature  is composed of 9 states:

\begin{equation}\label{eq:state_vector}
    \textbf{x} = [\tilde{x}, \tilde{y}, \tilde{\psi}, u,v,r, l,d_{o}, \tilde{\psi}_{o}],
\end{equation}

where $\tilde{x}$ and  $\tilde{y}$ represent the distance to the berthing point in the body frame, while $\tilde{\psi}$ represents the difference in the heading compared to the desired heading. The vessel velocities are given by the variables \textit{u, v}, and \textit{r.} The binary variable \textit{l} indicates whether or not the vessel has made contact (crashed) with an obstacle. The relative position to the closest obstacle in body frame is given by $d_{o}$ and $\tilde{\psi_{o}}$.

%\setlength{\textfloatsep}{2pt}
% trim={<left> <lower> <right> <upper>}

The PPO-trained policy network involves two hidden layers, consisting of 400 neurons each. The \textit{ReLU} activation function was used for the hidden layers, and the hyperbolic tangent was used as the activation function for the output layer, ensuring actions in the range [-1,1]. The PPO-trained policy converged after approximately 6 million interactions with the environment. The DRL agent was trained in   \cite{RorvikElla-LoviseHammervold2020Adae} to perform both the approach and berthing phase of docking, but without consideration for any speed regulations within the harbor.

\section{Approximation via Linear Model Trees}
A decision tree is a rule-based prediction method, which splits the input space into smaller regions and makes a prediction for each region \cite{MurphyKevinP.2012Ml:a}. A tree consists of branch nodes, where the splitting happens, and leaf nodes, where the prediction happens. LMTs perform linear regression in all the regions separately instead of attempting to fit the function for the entire feature space at once. To preserve the transparency of the tree in this work, the splits are univariate. Increasing the complexity of the prediction- or splitting model significantly increases the computational time required to build a tree, in addition to limiting their transparency. More generally, given any black-box model\textit{ f: x} $\rightarrow$\textit{ y}, LMTs make out a piece-wise linear approximation function \textit{f': x} $\rightarrow$ \textit{y'}, where $y  \simeq y'$. LMTs are useful because they are easy to implement and can give explanations in real-time, which is crucial for most robotic applications. The implementation of the LMTs in this paper is based on \cite{AnsonWongGithub} which again is based on Classification and Regression Trees (CART) from \cite{breiman_classification_1984}. The following modifications have been made to \cite{AnsonWongGithub}'s implementation:
\begin{itemize}
    \item We added randomization in the process of searching for thresholds and choosing the next node to split.
    \item We replaced the maximum depth of the tree with maximum number of leaf nodes.
\end{itemize} % fix, introduce normal way and then highlight the differences
The tree building process as implemented in this paper is shown in Algorithm \ref{alg:1}.

\begin{algorithm}[tb]
\SetAlgoLined
\textbf{Require: }\textit{\\training data $\mathcal{D}$
\\ Maximum number of leaf nodes \textbf{N}
\\ Minimum number of data samples for leaf nodes \textbf{M}}\\
\While{number of leaf nodes is less than \textbf{N}}{
\eIf{there exist a node that fulfills all splitting criteria}{
Choose node to split\\
Perform splitting \\
Calculate best potential split for the newly created nodes\\}{ \textbf{return} root node}
}\caption{Building LMTs}
\label{alg:1}
\end{algorithm}

For a node to be split, and two new nodes to be created, there must exist at least one node which fulfills the splitting criteria. That is, a node where a split will result in two nodes with at least \textbf{M} data samples each, and gives an improvement in the model's loss. When a splitting has occurred, the best potential split for the newly created nodes are calculated. It is this loss improvement value that is used when choosing the next node to be split. When searching for these split conditions, it is not possible to check for all possible thresholds, so instead a grid search is done. There is no guarantee that the optimal threshold will be found in this grid search, so some randomness is introduced. Not having the process be deterministic is beneficial because the process will generate a different tree every time it is run, which allows us to explore different outcomes. Equations \ref{eq:min_loss}-\ref{eq:thresholds} show how the split variables for a node are calculated.
\begin{equation}\label{eq:min_loss}
    f, t_n = \underset{f,n}{\mathrm{argmin}} ( loss (\mathcal{D}_L) + loss (\mathcal{D}_R))
\end{equation}

\begin{equation} \label{eq:split_cond}
\begin{split}
    \mathcal{D}_L &  = (x \in \mathcal{D} : x_f \leq t_n)\\
    \mathcal{D}_R & = (x \in \mathcal{D} : x_f > t_n)
\end{split}
\end{equation}

\begin{equation}\label{eq:thresholds}
    t_n = min (\mathcal{D}_f)+ (n+r)\frac{ (max (\mathcal{D}_f) - min (\mathcal{D}_f))}{N}
\end{equation}
where \textit{f} is the feature the split should be performed on, $t_n$ is the threshold number \textit{n} in the grid search, where $n \in [0,1,2,...,N] $, N is the size of the grid search, and \textit{r} is a random number between $\pm$2\%. The variable $\mathcal{D}_f$ denotes the values of feature \textit{f} in the set $\mathcal{D}$, thus \textit{min($\mathcal{D}_f$}) and \textit{max($\mathcal{D}_f$}) denotes the minimum and maximum values of feature \textit{f} in dataset $\mathcal{D}$. It is important to note that the LMT training process makes local, greedy choices, which gives no guarantee of global optimality. For example, if an extremely good split comes after an apparently bad split, it may never be found. This is a common problem for heuristic decision tree training methods. We alleviate this issue by adding some randomness to the process of choosing the next node to be split, as shown in the following equation:
% move equation 3 furthest down, fix text, add $ $ around variables
\begin{equation}\label{eq:split_chooser}
    n_{s} = \underset{n}{\mathrm{argmax}} ( (1+ r) (loss ( \mathcal{D}_L^{n}) + loss ( \mathcal{D}_R^{n})))
\end{equation}

where $n_s$ is the node to be split next, \textit{r} is a random number between $\pm$2\%,  and $\mathcal{D}_L^{n}$ and $\mathcal{D}_R^{n}$ are as defined in Equation \ref{eq:split_cond} given the best split conditions \textit{f} and \textit{t} for node n.

The tree has no maximum depth condition, instead it has a maximum number of leaf nodes it is allowed to have. This lets us directly state how many regions the tree is allowed to divide the input feature space into. Additionally, the tree is allowed to grow more asymmetrically, which again allows the tree to grow deeper in the area that covers the most complex regions of the feature space, while keeping the parts of the tree that covers simpler regions shallower.

The aspect that proved to be most important, and most challenging, was getting a balanced data set. The number of data points a certain area in the feature space requires in order to be represented adequately, depends on how far away from linearity the problem is in that area. To account for this, iterative tree-building was used. First, an initial data set is created through randomly sampling the feature space. Then, an initial tree is created based on that data set. The data set is then improved by letting the tree run through the environment and further samples from episodes that did not end successfully. To form the local explanations, the linear functions in the leaf nodes are utilized. The linear functions take the form of Equation \ref{eq:lin_leaf} where $a_f$ is feature \textit{f}'s coefficient and $x_f$ is the sample \textit{x}'s value for feature \textit{f}, and C is a constant. The importance $I_f$ for each input feature concerning each output feature can be calculated as shown in Equation \ref{eq:feature_importnace}, similarly to LIME and SHAP. 
\begin{equation}\label{eq:lin_leaf}
    y = \sum_f a_f x_f + C
\end{equation}
\begin{equation}\label{eq:feature_importnace}
    I_{f} = \frac{a_f x_f}{\sum_{j \in \forall f} | a_j x_j |}
\end{equation}
% decomposable transaprent? make clear that the tree can grow very big
Transparency can be divided into three categories, namely simulatable-, decomposable-, and algorithmic transparency. In \cite{BarredoArrietaAlejandro2020EAIX}, simulatable transparency is defined as the model not being more complex than what a human can easily comprehend. Therefore, given that the input features are understandable by humans (or at least domain experts) and the tree is not too deep, a linear model tree can be simulatable transparent. Decomposable transparency means that every part of the model must be understandable by a human without any additional tools. Since the linear model trees have univariate splits and linear function in the leaf nodes and the input and output features are understandable they are decomposable transparent, even if they grow big. Algorithmic transparency takes into account if it is possible to analyze the model with help of mathematical tools, which it is. Thus, linear models with univariate splits can be simulatable transparent but are always decomposable- and algorithmic transparent. The explanations given by the LMTs are \textit{local, feature relevant explanations}, which means that, for each prediction, an explanation in the form of showing how much a feature pulled in a certain direction is given. 

\begin{figure}[tb]
\begin{subfigure}{0.45\textwidth}

  % include first image
  \flushleft
  \includegraphics[width=\textwidth]{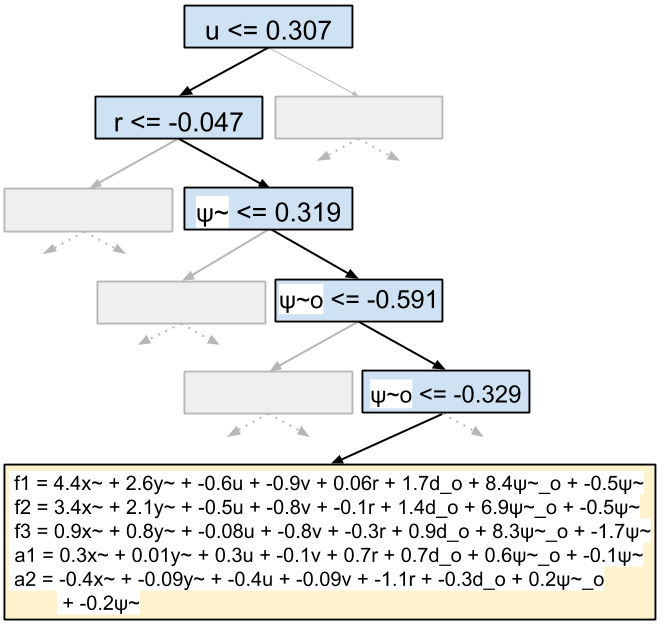}
  \captionsetup{justification=raggedright,singlelinecheck=false}
    \caption{Tree path for actions. Left arrow means that the condition in branch node above was true, right arrow means it was false.}
    \label{fig:lmt_ill}
\end{subfigure}
\\
\begin{subfigure}{0.45\textwidth}
\flushleft
  % include second image
  \includegraphics[width=\textwidth]{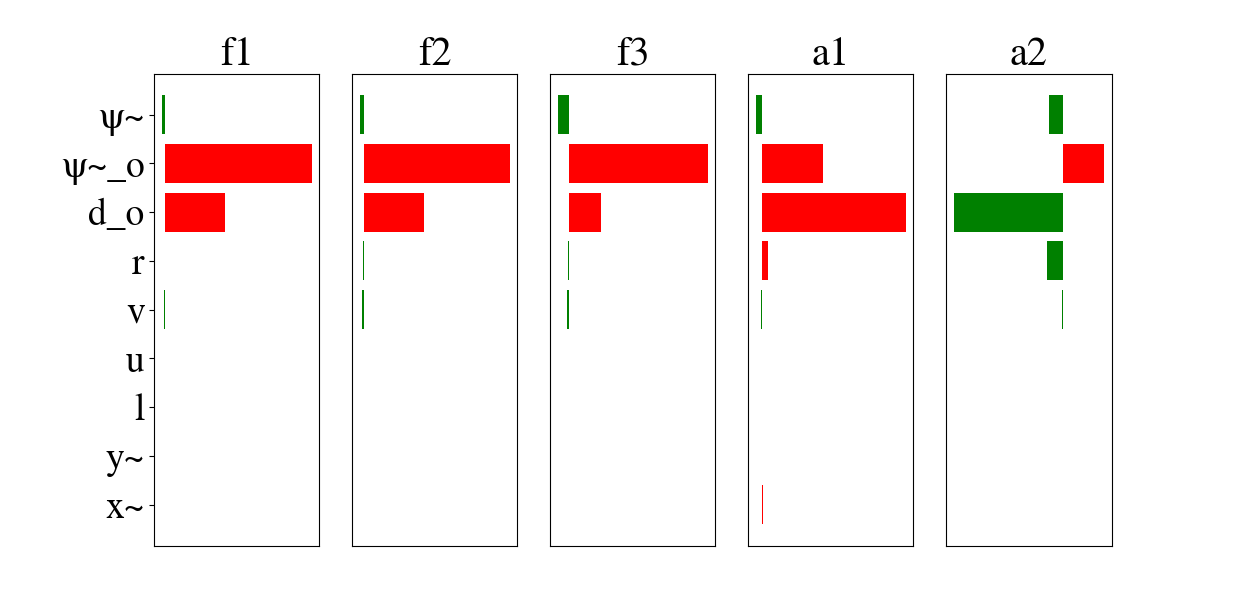}
  \captionsetup{justification=raggedright,singlelinecheck=false}
    \caption{Relative importance for input features for the actions}
    \label{fig:explantions}
\end{subfigure}
%\todo[inline]{when use action and when use output feature}
\caption{Explanations and path from root node to leaf node predicting the actions in last step of Figure \ref{fig:ppo_18}.}
\label{fig:pathsd_18}
\end{figure}
\setlength{\textfloatsep}{2pt}
\begin{figure*}[tb]
     \centering
    \includegraphics[trim={2cm 0 0 0.26cm},width=0.9\textwidth]{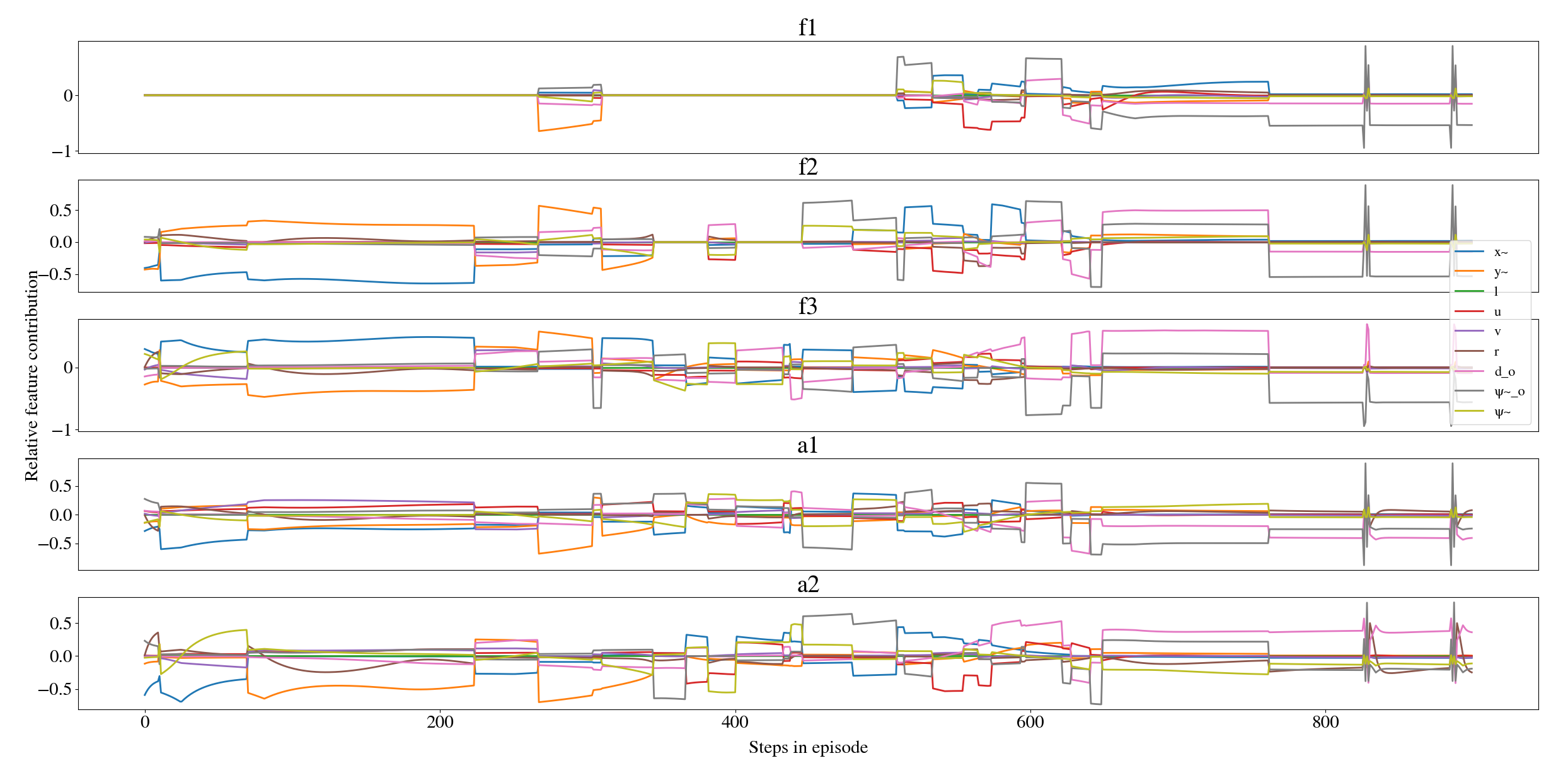}
    \caption{Relative feature contributions given by the LMT for the episode shown in Figure \ref{fig:ppo_18}.}
    \label{fig:expl_ppo_18}
\end{figure*}
%\subsection{Present structure of tree}
% trim={<left> <lower> <right> <upper>}

\section{Simulation results}\label{sec:exp_results}

%\todo[inline]{ppo-agent vs ppo-trianed policy}

\begin{figure}[]
\begin{subfigure}{.45\textwidth}
  %\flushleft
  % include first image
  \includegraphics[trim={8.2cm 0 8cm 2cm},clip,width=0.9\textwidth]{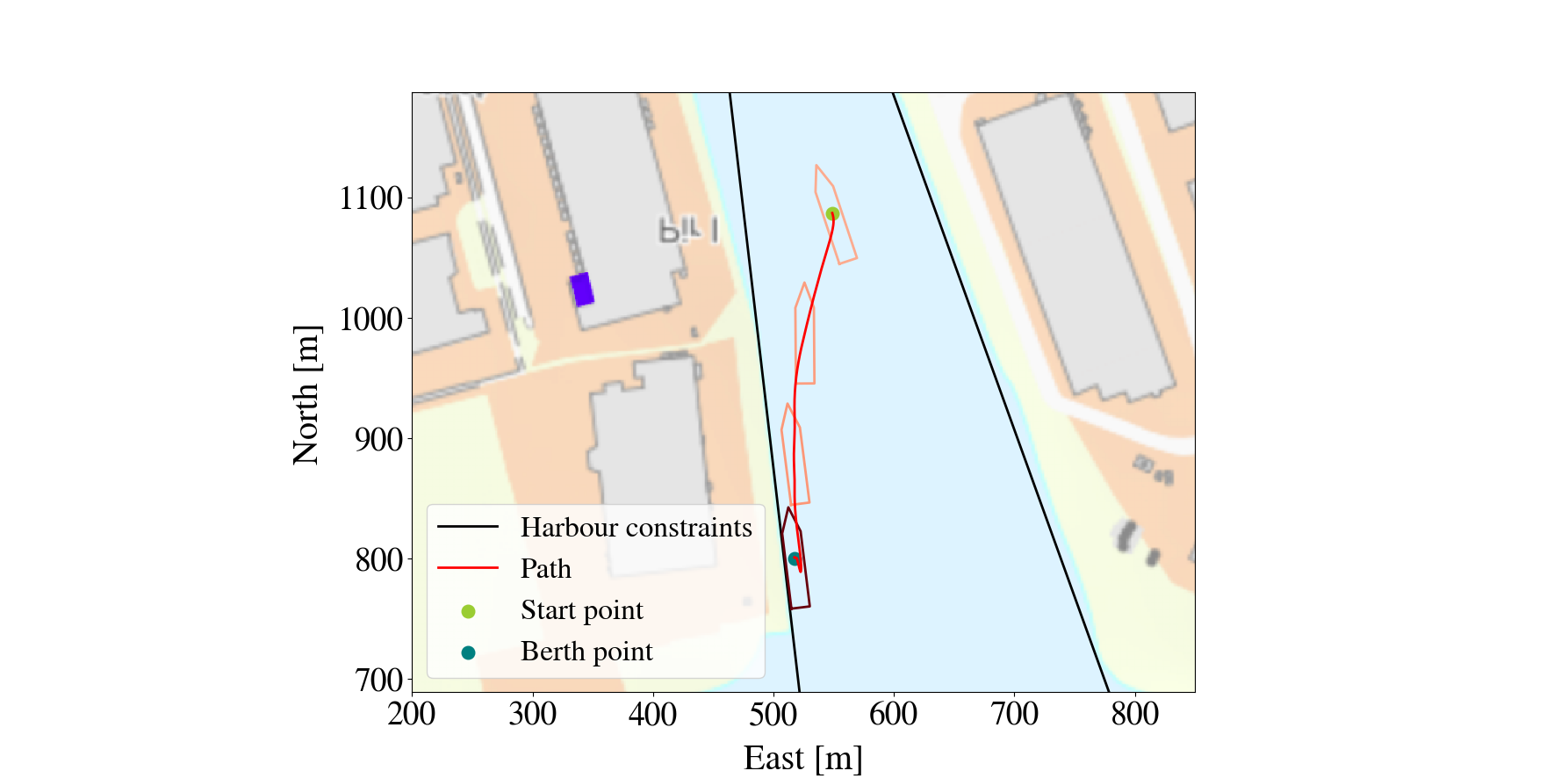}  
  \caption{Successful run by the LMT.}
  \label{fig:lmt_18}
\end{subfigure}
\\
\begin{subfigure}{.45\textwidth}
  %\flushright
  % include second image
  \includegraphics[trim={8.2cm 0 8cm 2.6cm},clip,width=0.9\textwidth]{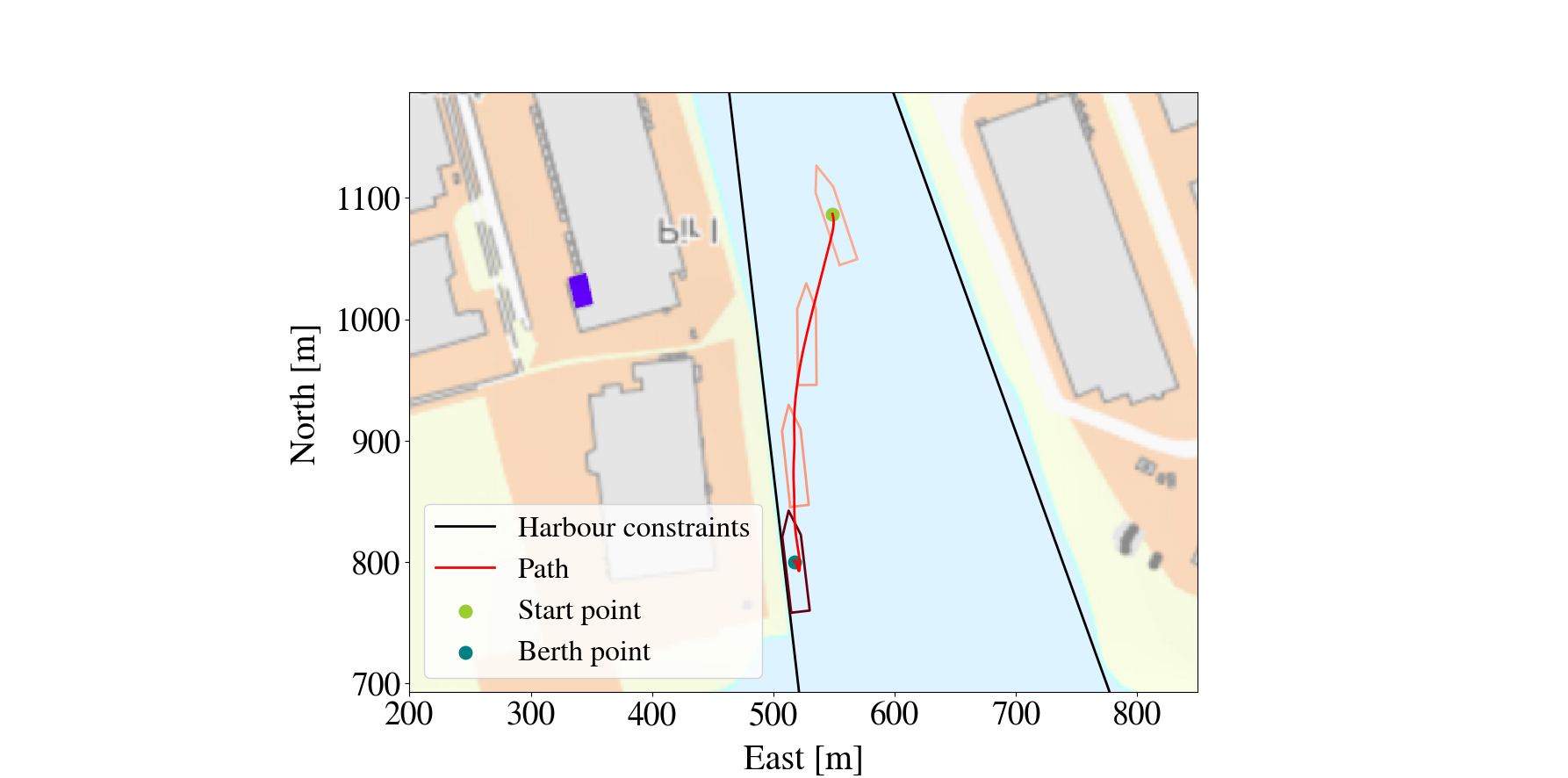}  
  \caption{Successful run by the PPO-agent.}
  \label{fig:ppo_18}
\end{subfigure}
\caption{Paths of PPO-agent and LMT from same starting point.}
\label{fig:paths_18}
\end{figure}

For this application, there are five control inputs to be predicted. This can either be solved by fitting one tree to each output feature or combining their losses when evaluating splits. The LMT made for this work used the latter and had 681 leaf nodes, where the shallowest leaf nodes were at depth 5, and the deepest leaf nodes were at depth 15. Thus, for all practical means, the resulting tree is only decomposable transparent, and not simulatable transparent. In Figure \ref{fig:lmt_ill}, the path along the tree from the root node to leaf node is highlighted, and in Figure \ref{fig:explantions} the explanations given in form of relative feature contribution is shown. Figure \ref{fig:pathsd_18} is from the last time instance shown in Figure \ref{fig:ppo_18}. Like the PPO-agent, the LMT can act as a controller on its own. Thus, how well the LMT approximated the PPO-trained policy network can be evaluated through the difference between their outputs when they are given the same input. Table \ref{tab:error} shows the analysis of the LMT's error through its deviation from the target output by the PPO-agent from 1000 episodes with random starting points. In most episodes, the vessel has arrived at the berthing point at step 800. After this it enters a cycle of repeating states. To prevent these states to skew the evaluation since the LMT approximates the PPO-agent quite well in the region close to the berthing point, the episodes are stopped at step 800. The highest errors usually occurs in the beginning of the episode, when the vessel still is far from the harbour, where the LMT's actions follows the curvature of the PPO-agent's actions but are somewhat noisy, causing an increase in the average error. Overall, the magnitude of the average error and standard deviation is moderate.

% Add the range of outputs, so that this means something

\begin{table}[tb]
    \centering
    \begin{tabular}{|c||c|c|}
\hline
   \textbf{ Output feature} & \textbf{Mean absolute error} & \textbf{Error standard deviation} \\ \hline \hline
   \textbf{$f_1$}( kN ) & 15.84(9.3\%) & 25.6(15.05\%) \\ \hline
    \textbf{$f_2$}( kN )& 14.23(8.3\%) & 21.7(12.76\%) \\ \hline
   \textbf{ $\alpha_1$}( deg ) & 16.61(9.2\%) & 23.49(13.05\%)  \\ \hline
    \textbf{$\alpha_2$}( deg ) & 13.75(7.63\%) & 20.62(11.45\%)\\ \hline
    \textbf{$f_3$}( kN ) & 9.08(9.08\%) & 15.9(15.9\%)\\ \hline
    
\end{tabular}
\caption{Output error analysis}
\setlength{\textfloatsep}{2pt}
\label{tab:error}

\end{table}

%\todo[inline]{bigger titles and x title is gone?}

\begin{figure}[tb]
\begin{subfigure}{.5\textwidth}
  \flushleft
  % include first image
  \includegraphics[trim={4cm 0.25cm 4cm 2cm},clip,width=0.9\textwidth]{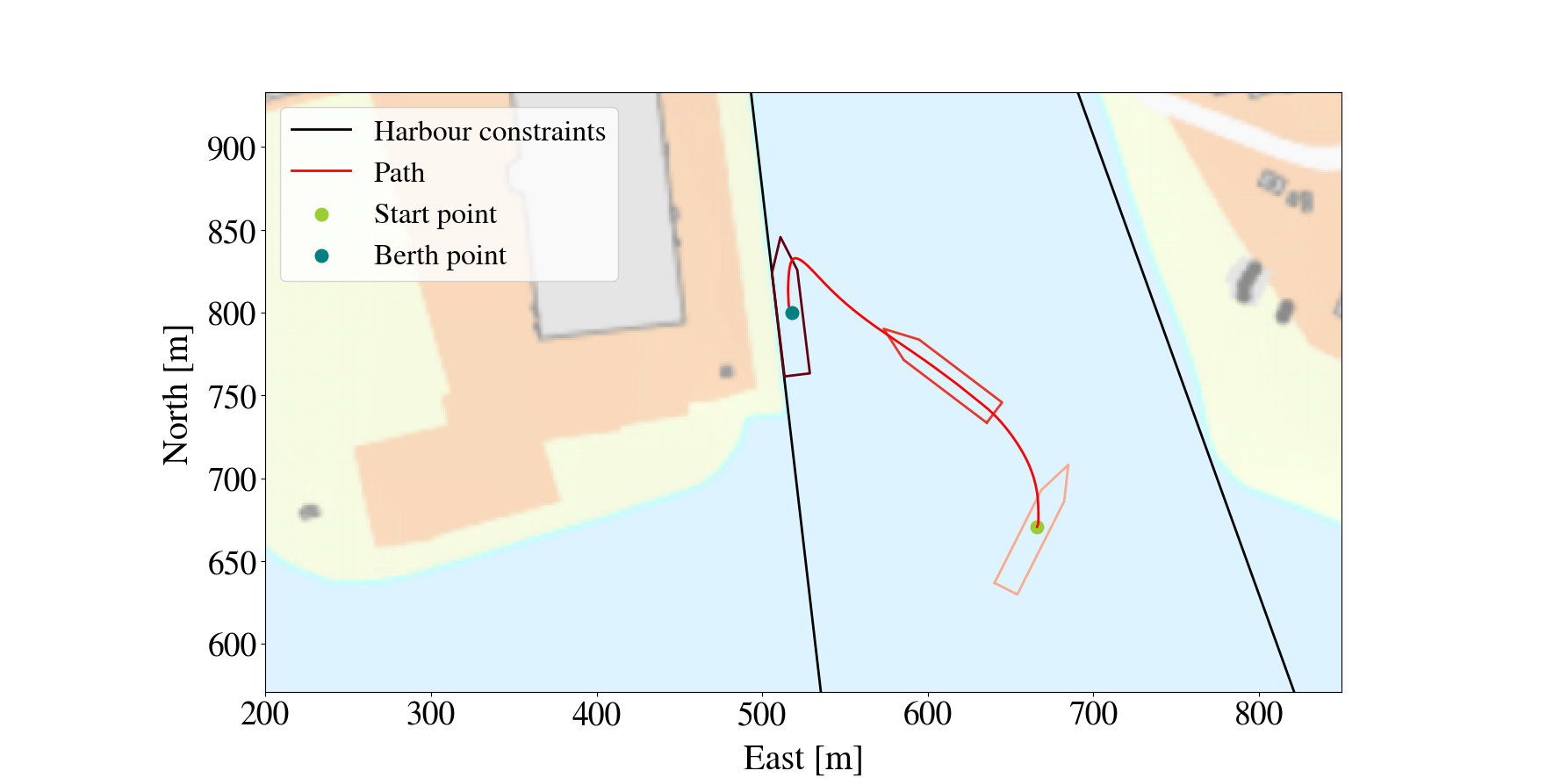}  
  \caption{Failed run by the LMT.}
  \label{fig:lmt_25}
\end{subfigure}
\\
\begin{subfigure}{.5\textwidth}
  \flushright
  % include second image
  \includegraphics[trim={4cm 0.25cm 4cm 2.6cm},clip,width=0.9\textwidth]{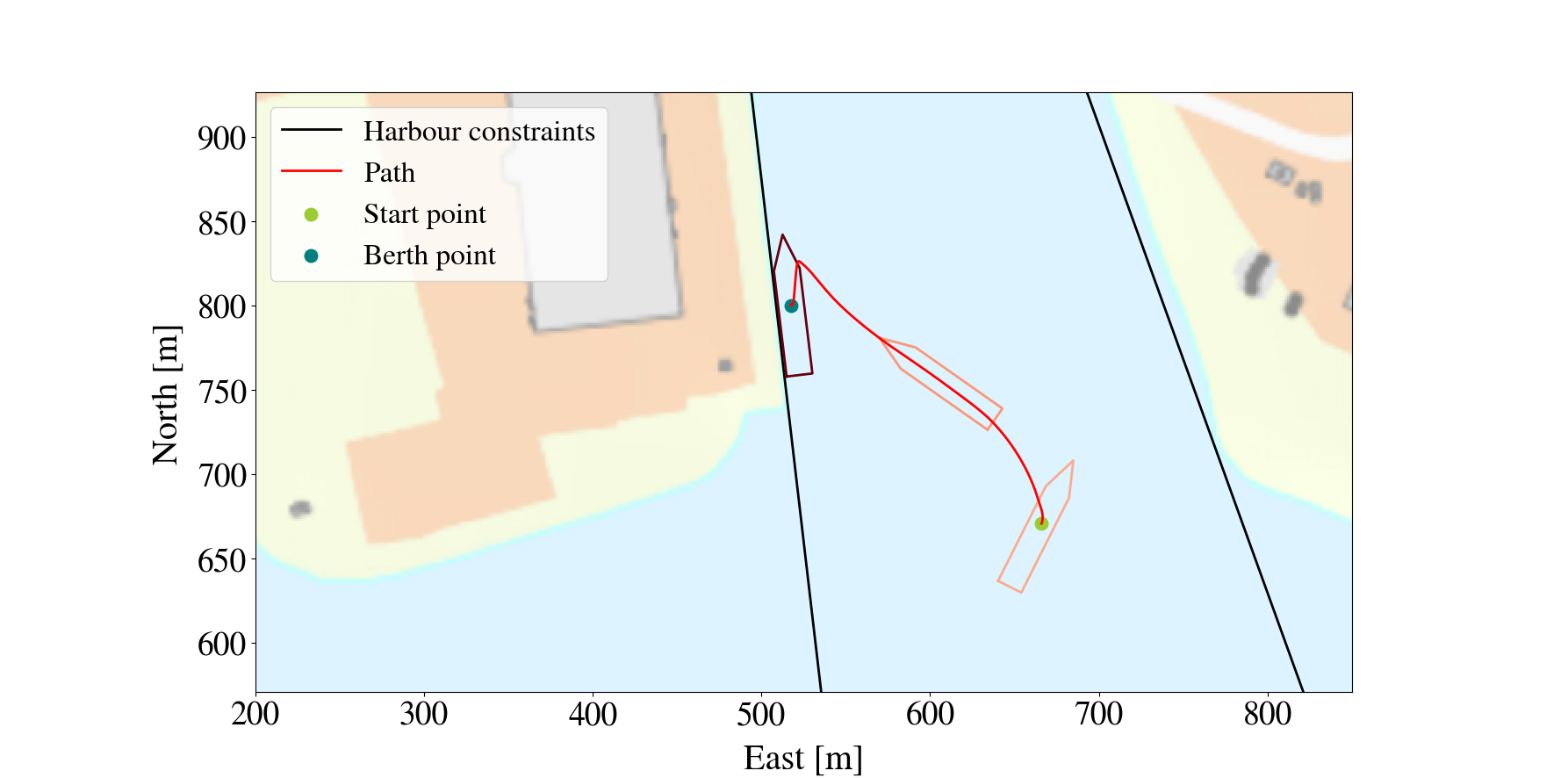}  
  \caption{Successful run by the PPO-agent.}
  \label{fig:ppo_25}
\end{subfigure}
\caption{Paths of PPO-agent and LMT from same starting point.}
\label{fig:paths_25}
\end{figure}

\begin{figure}[!htp]
\begin{subfigure}{.48\textwidth}
  \flushleft
  % include first image
  \includegraphics[trim={2.7cm 1.8cm 3.57cm 3.95cm},clip,width=0.9\textwidth]{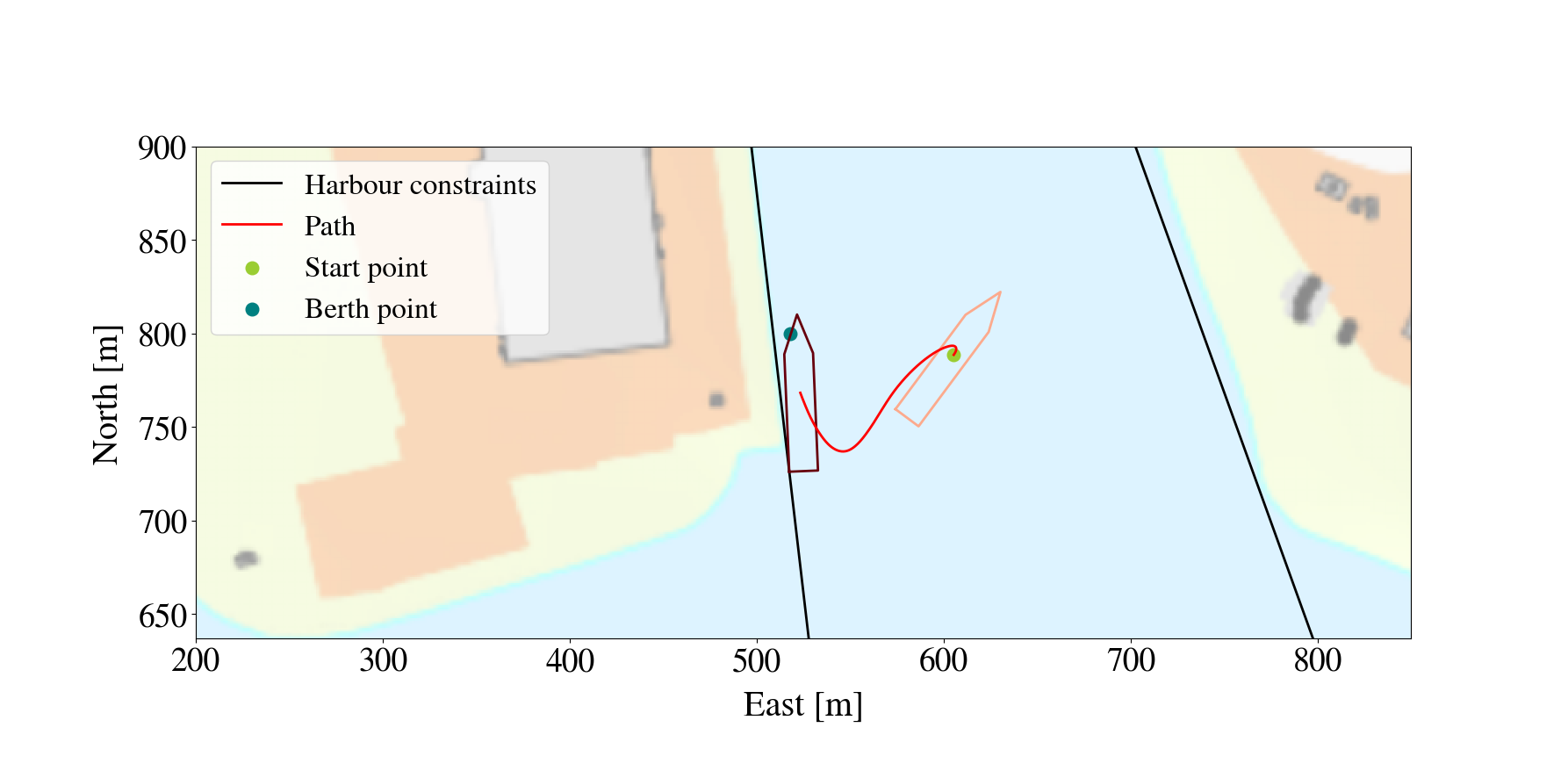}  
  \caption{Failed run by the LMT.}
  \label{fig:lmt_23}
\end{subfigure}
\\
\begin{subfigure}{.48\textwidth}
  \flushleft
  % include second image
  \includegraphics[trim={2.7cm 1.9cm 3.57cm 3.95cm},clip,width=0.9\textwidth]{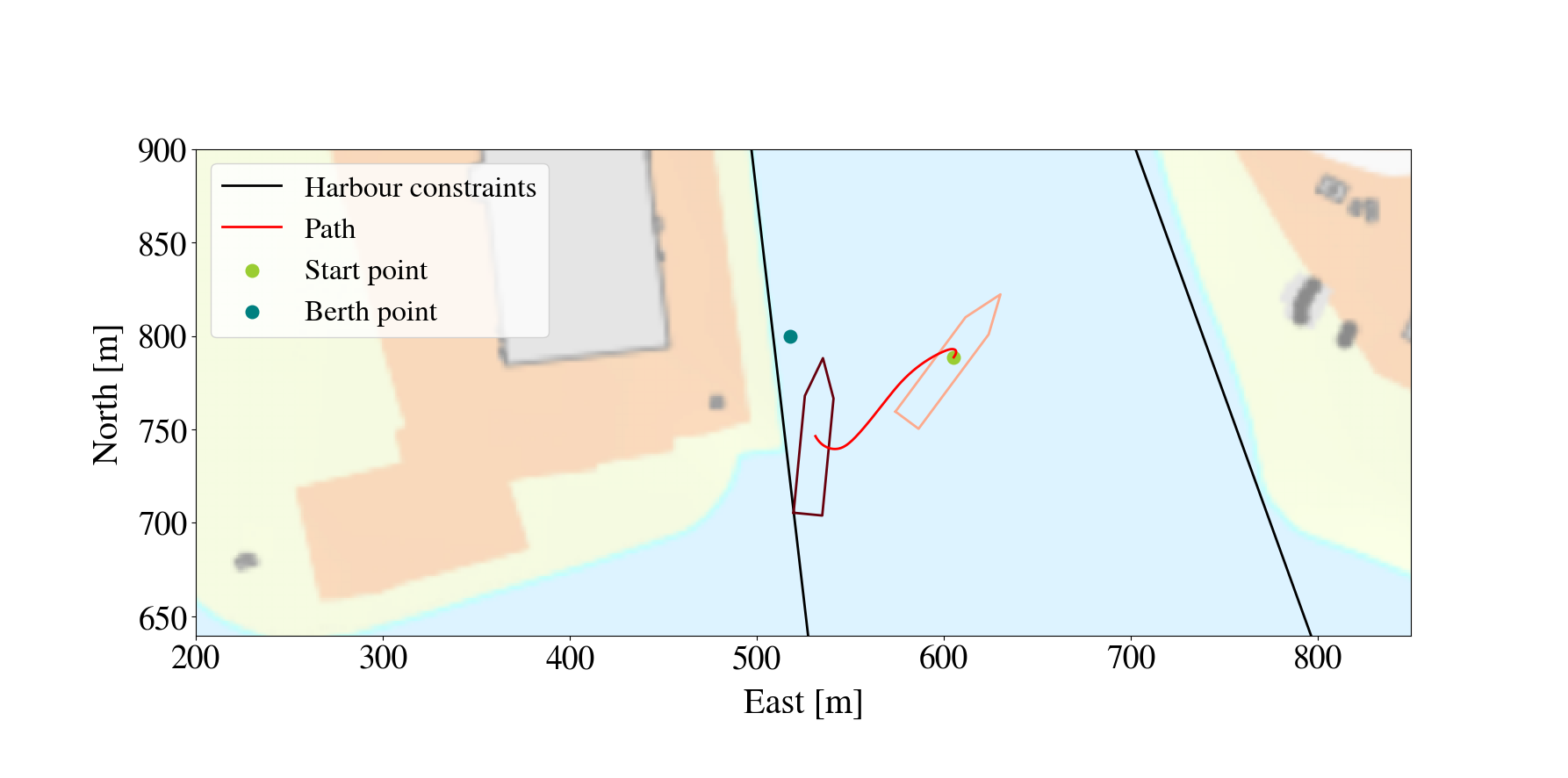}  
  \caption{Failed run by the PPO-agent.}
  \label{fig:ppo_23}
\end{subfigure}
\caption{Paths of PPO-agent and LMT from same starting point.}
\label{fig:paths_23}
\end{figure}

An alternative way to evaluate the LMT's approximation of the PPO-trained DRL agent is to look at their paths when starting from the same initial point and aiming for the same berthing point. A successful run is here defined as the vessel reaching the berthing point without making contact with any obstacle, while a failed run is defined as the vessel making contact with an obstacle (crashing). This criterion is not meant to evaluate the PPO-agent's behavior, because berthing can be successful even if it makes contact with the harbor if it happens slowly enough (i.e a small bump is usually tolerated), but rather as a way of evaluating how well the LMT managed to approximate the PPO-trained policy. Of note, neither of the agents have any episodes that does not end by either successfully berthing the vessel or by making contact with an obstacle. An example of a successful run by both the LMT and the PPO-agent is shown in Figure \ref{fig:lmt_18}. It is clear that for this starting point, the LMT has approximated the PPO-trained policy very well. The LMT fails approximately 3\% more often than the PPO-agent, but when looking closer at the situations where the LMT fails and the PPO-agent succeeds, it is apparent that such episodes typically unfolds similar to the episode shown in Figure \ref{fig:paths_25}.  Even though their outcome is different, they act very similarly, so the explanations are still useful. However, the biggest difference between the two agents is most apparent when the PPO-agent fails, as can be seen in Figure \ref{fig:paths_23}. This could either be due to the LMT not having seen enough data from this area, that this is a more complex area so the LMT needs to grow deeper, or that the PPO-agent has not found a proper strategy for this area (which in turn can be due to the starting position being extremely hard or even impossible, for example, if the boat has an initial speed towards the harbor that is too high. However, if this deviation is detected, it might be used to raise an alarm of some sort, to alert an overseer. The explanations for the episode shown in Figure \ref{fig:ppo_18} are shown in Figure \ref{fig:expl_ppo_18}. Since the LMT only uses the linear functions in the leaf nodes as a basis for its explanations it does not take the splits along the path from the root node to the respective leaf node into consideration, even though it intuitively is relevant. This can for example be seen in the two flat areas in the first 500 steps of the episode for output $f_1$. The PPO-agent reaches the berthing point at around step number 750, and both the output of the PPO-agent and the explanations from the LMT goes into a rather repetitive cycle. LMT assigns most importance to $\tilde{\psi}_o$ and $d_o$ for all actions. When looking closer at what features are changing in this part of the episode, it is clear that it is in fact $\tilde{\psi}_o$, \textit{r}, and $d_o$ that are changing the most, while $\tilde{x}$ and $\tilde{y}$ are virtually constant. Since feed-forward neural networks are one-to-one, the changing parameters are causing the change in outputs and are therefore the correct explanations. In the first $\sim$ 250 steps it seems like the PPO-agent cares most about the three input features regarding the vessel's position relative to the berthing point ($\tilde{x},  \tilde{y}$ and $\tilde{\psi}$). The part where the explanations are the least decisive is from approximately step number 250 to 750. LMTs explanations changes fast, which reflects that it is only piece-wise smooth. LMTs are somewhat time-demanding to build, but when it is built they can easily give real-time explanations. The LMTs only give explanations for one output feature at a time. The problem with this is that the 5 outputs are controlling the same vessel and thus dependent on each other. Additionally, $f_1$ and $\alpha_1$, and $f_2$ and $\alpha_2$ are controlling the same motor. Explaining dependent factors independently will not give the whole picture. Even though relative feature contributions cannot serve as a full-fledged explanation in itself, it can be an important component of technical assurance\cite{Glomsrud19}.

% trim={<left> <lower> <right> <upper>}
\setlength{\textfloatsep}{2pt}
% trim={<left> <lower> <right> <upper>}
\setlength{\textfloatsep}{2pt}
% trim={<left> <lower> <right> <upper>}
\setlength{\textfloatsep}{2pt}

\newpage

\section{Conclusions and future work}
Linear model trees (LMTs) can contribute to tracing the outputs of a deep reinforcement learning (DRL) policy by directly linking them to the input features. In this paper, this potential was demonstrated by approximating a DRL policy controlling an autonomous surface vessel with five control inputs in a complex motion control scenario, namely docking. Although LMTs do not approximate the deep neural network in an optimal way, our results indicate that their performance is close enough to that of the original policy. In addition, the fact that LMTs are fast enough to be applicable in real-time applications, make them good candidates as components a digital assurance framework explaining the actions of black box models during operation. Future work includes improving the accuracy of the trees, and utilizing domain knowledge in both the process of building the trees and the process of extracting information about the system from the trees to make them truly understandable to several categories of end users.

%This is a step towards utilizing data-driven methods to the fullest in cost- and safety-critical applications.
% In  thispaper,  the  main  contribution  is  the  use  of  a  linear  model  tree(LMT)  to  approximate  a  DNN  policy,  originally  trained  viaproximal policy optimization, for an autonomous surface vehicleperforming a docking operation. The two main benefits of theproposed approach are: a) LMTs are transparent and make itpossible to associate directly the outputs (control actions, in ourcase) with specific values of the input features, b) LMTs are veryfast  and  can  provide  information  in  real-time.  Our  simulatedresults indicate that LMTs can be a useful component, alongsideothers,  within  a  digital  assurance  framework

%\clearpage
\FloatBarrier
\bibliography{bib}
\bibliographystyle{ieeetr.bst}
\vspace{12pt}

\end{document}